\documentclass[runningheads]{llncs}
\usepackage{amsmath,amssymb} % define this before the line numbering.
\usepackage{color}
\usepackage{graphicx}
\usepackage{epsfig}
\usepackage{subfigure}
\usepackage{epstopdf}
\usepackage{enumerate}
\usepackage{booktabs}
\usepackage{empheq}
\usepackage{bbding}
\usepackage{cite}
\usepackage{ulem}
\usepackage{booktabs}
\usepackage[american]{babel}
\usepackage{url}

\begin{document}
\title{Joint Representation and Truncated Inference Learning for Correlation Filter based Tracking}
% Replace with your title

\titlerunning{Joint Representation and Truncated Inference Learning}
% Replace with a meaningful short version of your title
%
\author{Yingjie Yao\inst{1}\orcidID{0000-0002-3533-1569} \and
Xiaohe Wu\inst{1}\orcidID{0000-0001-6884-9121} \and
Lei~Zhang\inst{2}\orcidID{0000-0002-4424-4942} \and
Shiguang Shan\inst{3}\orcidID{0000-0002-8348-392X} \and
Wangmeng Zuo\inst{1} \inst{(}\Envelope\inst{)} \orcidID{0000-0002-3330-783X}
}
%
%Please write out author names in full in the paper, i.e. full given and family names.
%If any authors have names that can be parsed into FirstName LastName in multiple ways, please include the correct parsing, in a comment to the volume editors:
%\index{Lastnames, Firstnames}
%(Do not uncomment it, because you may introduce extra index items if you do that, we will use scripts for introducing index entries...)
\authorrunning{Yao et al.}
% Replace with shorter version of the author list. If there are more authors than fits a line, please use A. Author et al.
%

\institute{Harbin Institute of Technology, Harbin 150001, China \and
University of Pittsburgh, 3362 Fifth Avenue, Pittsburgh, PA 15213 \and
Institute of Computing Technology, CAS, Beijing, 100049, China
\email{\{yaoyoyogurt,xhwu.cpsl.hit,cszhanglei\}@gmail.com,\\
sgshan@ict.ac.cn, wmzuo@hit.edu.cn}
}
\maketitle              % typeset the header of the contribution
\begin{abstract}
Correlation filter (CF) based trackers generally include two modules,
i.e., feature representation and on-line model adaptation.
In existing off-line deep learning models for CF trackers,
the model adaptation usually is either abandoned or has closed-form solution
to make it feasible to learn deep representation in an end-to-end manner.
However, such solutions fail to exploit the advances in CF models,
and cannot achieve competitive accuracy in comparison with the state-of-the-art CF trackers.
In this paper, we investigate the joint learning of deep representation and model adaptation,
where an updater network is introduced for better tracking on future frame
by taking current frame representation, tracking result, and last CF tracker as input.
By modeling the representor as convolutional neural network (CNN),
we truncate the alternating direction method of multipliers (ADMM)
and interpret it as a deep network of updater, resulting in our model for learning representation and truncated inference (RTINet).
Experiments demonstrate that our RTINet tracker achieves favorable tracking accuracy against the state-of-the-art trackers
and its rapid version can run at a real-time speed of 24 fps.
The code and pre-trained models will be publicly available at \url{https://github.com/tourmaline612/RTINet}.

\end{abstract}
%
%
%
% -----------------------------------------------------------------------------------------------------
% -----------------------------------------------------------------------------------------------------
\section{Introduction}
In recent years, correlation filters (CFs) have achieved noteworthy advances as well as state-of-the-art performance in visual tracking.
Generally, the CF-based approaches learn CFs on feature representation
for model adaptation along with an image sequence.
Therefore, the advancement of CF-based tracking performance is mainly
driven by the improvement on both feature representation and CF learning model.
The development of feature representation has witnessed the evolution
from handcrafted HOG~\cite{henriques2015high} and ColorNames (CN)~\cite{danelljan2014adaptive}
to deep convolutional neural network (CNN) features~\cite{ma2015hierarchical,danelljan2015convolutional,qi2016hedged}.
And their combination has also been adopted \cite{danelljan2017eco,CCOT}.
Meanwhile, the learning models have also been continuously improved with the introduction of spatial regularization~\cite{SRDCF,danelljan2015convolutional,danelljan2016adaptive},
continuous convolution~\cite{CCOT}, target response adaptation~\cite{bibi2016target},
context regularization~\cite{mueller2017context}, {temporal regularization~\cite{li2018STRCF},
and other sophisticated learning models~\cite{danelljan2017eco,BACF,zuo2018learning}}.

Motivated by the unprecedented success of CNNs~\cite{krizhevsky2012imagenet,vggnet,he2016deep,ren2015faster} in computer vision, it is encouraging to study the off-line training of deep CNNs for feature representation and model adaptation in CF trackers.
Unfortunately, model adaptation in CF tracking usually requires to solve a complex optimization problem, and is not trivial to be off-line trained together with deep representation.
To enable off-line training of deep representation specified for visual tracking, the Siamese network solutions~\cite{SiameseFC,tao2016siamese,chen2017once} are suggested to bypass the model adaptation by learning a matcher to discriminate whether a patch is matched with the exemplar image annotated in the first frame.
In~\cite{SiameseFC,tao2016siamese,chen2017once}, the tracker is fixed since the first frame, and cannot adapt to the appearance temporal variation of target.
For joint off-training of deep representation and model adaptation, Valmadre et al.~\cite{CFNet} adopt the original CF form due to its model adaptation has the closed-form solution and can be interpreted as a differentiable CNN layer.
Instead of directly taking model adaptation into account, Guo et al. \cite{guo2017learning} suggest a dynamic Siamese network for modeling temporal variation, while Choi et al. \cite{choi2017deep} exploit the forward-pass of meta-learner network to provide new appearance information to Siamese network.
These approaches, however, fail to exploit the continuous improvement on CF models~\cite{SRDCF,BACF,danelljan2015convolutional,CCOT}, and even may not achieve comparable tracking accuracy with the deployment of advanced CF models on deep features pre-trained for classification and detection tasks.

In response to the aforementioned issues, this paper presents a bi-level optimization formulation as well as a RTINet architecture for joint off-line learning of deep representation and model adaptation in CF-based tracking.
To exploit the advances in CF tracking, the lower-level task adopts a more sophisticated CF model~\cite{BACF} by incorporating background-aware modeling, which can learn CFs with limited boundary effect from large spatial supports.
And we define the upper-level objective on future frame for task-driven learning and improving the tracking accuracy.
With unrolled optimization, we truncate the alternating direction method of multipliers (ADMM) for solving the lower-level task to form our RTINet, which can be interpreted as an updater network based on the deep representation provided by another representor network.
Therefore, our RTINet model enables the end-to-end off-line training of both deep representation and truncated inference.
Furthermore, task-driven learning of truncated inference is also helpful in improving the effectiveness of the baseline CF tracker~\cite{CFNet}.
Experiments show that combining CNN with advanced CF tracker can benefit tracking performance, and the joint learning of deep representation and truncated inference also improves tracking accuracy.
In comparison with state-of-the-art trackers, our RTINet tracker achieves favorable tracking accuracy,
and its rapid version can achieve a real time speed of 24 fps.

To sum up, the contribution of this work is three-fold:
\begin{enumerate}
\item We present a framework, i.e., RTINet, for off-line training of deep representation and model adaptation.
    Instead of combining CNN with the standard CF tracker~\cite{CFNet}, we show that the combination with the advanced CF tracker (i.e., BACF~\cite{BACF}) can improve the tracking performance with a large margin.
\item The model adaptation of the advanced CFs generally requires to solve a complex optimization problem, making it difficult to jointly train the representor and updater networks.
    To tackle this issue, we design the updater network by unrolling the ADMM algorithm, and define the loss on future frame to guide the model learning.
          %Our key contribution is introducing a meta-learning learning method which can select optimal value of the hyper-parameters using off-line training on multiple future frame simultanousely.
\item Experiments show that our RTINet achieves favorable accuracy against state-of-the-art trackers, while its rapid version can perform at real time speed.
%\item In addition, we introduced a novel spatial regularization layer which can greatly reduce the boundary effect and can be seamlessly integrated into any CNN architecture.
\end{enumerate}
% -----------------------------------------------------------------------------------------------------
% -----------------------------------------------------------------------------------------------------
\section{Related Work}
\label{sec:related_work}
Deep CNNs have demonstrated excellent performance in many challenging vision tasks \cite{ren2015faster,dong2014learning},
and inspire numerous works to adopt deep features in CF based trackers \cite{danelljan2017eco, ma2015hierarchical, danelljan2015convolutional}.
These methods simply use the feature representation generated by CNNs pre-trained for image classification,
which, however, are not tailored to visual tracking.
Several Siamese networks, e.g., SINT \cite{tao2016siamese}, GOTURN \cite{held2016learning}, and SiameseFC \cite{SiameseFC},
have been exploited for the off-line learning of CNN feature extractor for tracking,
but both the feature extractor and tracker are fixed for the first frame,
making them generally perform inferior to state-of-the-arts.

As a remedy, Guo et al. \cite{guo2017learning} and Choi et al. \cite{choi2017deep} learn to on-line update
the feature extractor for adapting to appearance variation during tracking.
Instead of learning to update the feature extractor, Valmadre et al. \cite{CFNet} adopt the simple CF model to off-line learn deep representation.
Due to that the original CF has the closed-form solution, it can be interpreted as a differentiable CNN layer
and enables the joint learning of deep representation and model adaptation.
These aforementioned approaches fail to exploit the continuous improvement on CF models ~\cite{SRDCF,BACF,danelljan2015convolutional,CCOT},
and cannot compete with the advanced CF models based on deep features.

Another related work is the meta-tracker by Park et al. \cite{park2018meta} which automatically learns
fast gradient directions for online model adaptation of an existing tracker (e.g., MDNet~\cite{MDNet}).
In contrast, our RTINet focuses on the joint off-line learning of deep representation and
model adaptation in CF-based tracking.
Moreover, most advanced CF trackers are formulated as constrained optimization,
which cannot be readily solved by gradient descent as meta-tracker \cite{park2018meta} does.
Therefore, we truncate the ADMM algorithm for solving BACF~\cite{BACF,CCOT} to design the updater network,
and then present our RTINet that enables the end-to-end off-line training of both deep representation and truncated inference.
Furthermore, off-line learning of truncated inference also benefits the improvement on effectiveness of the baseline optimization algorithm~\cite{zuo2015discriminative,yan2016deep}.
%

% -----------------------------------------------------------------------------------------------------
% -----------------------------------------------------------------------------------------------------
\section{Proposed Method} \label{sec:method}
In this section, we present our RTINet approach for joint off-line training of
deep representation and model adaptation in CF trackers.
To this end, we first briefly revisit a recent CF tracker, i.e., BACF~\cite{BACF},
to deliver some insights, and then introduce the formulation,
network architecture, and learning of our RTINet.

\subsection{Revisiting BACF}
Let $\mathbf{z}_t \in \mathbb{R}^{m \times n \times L}$
and $\mathbf{f}_t$ denote the feature representation of the current frame $\mathbf{x}_t$,
and the CFs adopted at frame $t$, respectively.
In CF based trackers, tracking can be performed by
first computing the response map $\sum_{l=1}^L \mathbf{z}_{t,l} \star \mathbf{f}_{t,l}$
as the cross-correlation between $\mathbf{z}_t$ and $\mathbf{f}_t$,
and then locating the target based on the maximum of the response map.
Here, $\star$ denotes the convolution operator, and the cross-correlation
can be efficiently performed with the Fast Fourier Transform (FFT),
making CFs very encouraging and intensively studied in visual tracking.
The original CF model updates the CFs by solving the following problem,
\begin{equation}\label{eq:standard_cf}
\min_{\mathbf{f}} \frac{1}{2} \left\|\mathbf{y}_t - \sum_{l=1}^L\mathbf{z}_{t,l} \star \mathbf{f}_l \right\| ^2 +
 \frac{\lambda}{2} \sum_{l=1}^L \|\mathbf{f}_l\|^2,
\end{equation}
where $\mathbf{y}_t$ is a Gaussian shaped function based on the tracking result at frame $t$,
and $\lambda$ is the regularization parameter.

Recently, many advanced CF models have been suggested to improve the original CF, resulting in continuous performance improvement on visual tracking.
Here we take BACF~\cite{BACF} as an example, which learns CFs by better exploiting
real negative samples via background-aware modeling.
The BACF model can be equivalently formulated as,
\begin{equation}\label{eq:bacf}
\min_{\mathbf{f}, \mathbf{h}} \frac{1}{2} \left\|\mathbf{y}_t - \sum_{l=1}^L\mathbf{z}_{t,l} \star \mathbf{f}_l \right\| ^2 + \frac{\lambda}{2} \|\mathbf{h}\|^2, \mbox{ s.t. } \mathbf{f}_l = \mathbf{M}^{\top} \mathbf{h}_l,
\end{equation}
where $\mathbf{M}$ is a binary selection matrix to crop the center patch of an image.
The BACF model can be efficiently solved using the Alternating Direction Method of Multipliers (ADMM).
Accordingly, the augmented Lagrangian function of Eqn.~(\ref{eq:bacf}) can be expressed as,
\begin{equation}\label{eq:al_bacf}
\small
L(\mathbf{f}, \mathbf{h}, \boldsymbol{\mu}) \!= \frac{1}{2} \! \left\|\mathbf{y}_t \!-\! \sum_{l=1}^L\mathbf{z}_{t,l} \!\star\! \mathbf{f}_l \right\| ^2 \!+\! \frac{\lambda}{2} \|\mathbf{h}\|^2 \!+\! \sum_{l=1}^L \boldsymbol{\mu}_l^{\top} (\mathbf{f}_l \!-\! \mathbf{M}^{\top} \mathbf{h}_l) \!+\! \frac{\rho}{2} \sum_{l=1}^L \|\mathbf{f}_l \!-\! \mathbf{M}^{\top} \mathbf{h}_l\|^2 \,,
\end{equation}
where $\boldsymbol{\mu}$ denotes the Lagrange multiplier, and $\rho$ is the penalty parameter.
By introducing $\mathbf{g} = \frac{1}{\rho} \boldsymbol{\mu}$, the optimization on $\{\mathbf{f}, \mathbf{h}\}$ of Eqn. (\ref{eq:al_bacf}) can be equivalently formed as,
\begin{equation}\label{eq:al2_bacf}
L(\mathbf{f}, \mathbf{h}, \mathbf{g}) \!=\frac{1}{2} \! \left\|\mathbf{y}_t \!-\! \sum_{l=1}^L\mathbf{z}_{t,l} \!\star\! \mathbf{f}_l \right\| ^2 \!+\! \frac{\lambda}{2} \|\mathbf{h}\|^2 \!+\! \frac{\rho}{2} \sum_{l=1}^L \|\mathbf{f}_l \!-\! \mathbf{M}^{\top} \mathbf{h}_l + \mathbf{g}_l\|^2 \, .
\end{equation}
The ADMM algorithm can then be applied to alternatingly update $\mathbf{h}$, $\mathbf{g}$ and $\mathbf{f}$,
\begin{eqnarray} \label{eq:admm}
 \begin{cases}
{{\mathbf{h}}^{\left( k+1 \right)}} = \arg \underset{\mathbf{h}} {\mathop{\min }}\, \frac{\lambda}{2} {{{\left\| \mathbf{h} \right\|}^{2}} +} \frac{\rho}{2} \sum_{l=1}^L \|\mathbf{f}_l^{(k)} \!-\! \mathbf{M}^{\top} \mathbf{h}_l + \mathbf{g}^{(k)}_l\|^2  \\
{{\mathbf{g}}_l^{\left( k+1 \right)}} = {{\mathbf{g}}_l^{\left( k \right)}} + {{\mathbf{f}}_l^{\left( k \right)}} - \mathbf{M}^{\top} {{\mathbf{h}}_l^{\left( k+1 \right)}} \\
{{\mathbf{f}}^{\left( k+1 \right)}} = \arg \underset{\mathbf{f}}{\mathop{\min }}\,\frac{1}{2}{{\left\| \mathbf{y}_t - \sum\limits_{l=1}^{L}{{{\mathbf{z}}_{t,l}} \star {{\mathbf{f}}_{l}}} \right\|}^{2}} +
\frac{\rho}{2} \sum_{l=1}^L \|\mathbf{f}_l \!-\! \mathbf{M}^{\top} \mathbf{h}^{(k+1)}_l + \mathbf{g}^{(k+1)}_l\|^2   \\
\end{cases}
\end{eqnarray}
We note that the subproblems on ${{\mathbf{f}}^{\left( k+1 \right)}}$ and ${{\mathbf{h}}^{\left( k+1 \right)}}$ have
closed-form solutions.
Once the solution $\mathbf{f}^*$ to Eqn. (\ref{eq:bacf}) is obtained, the CFs adopted at frame $t+1$ can then be attained with the linear interpolation updating rule defined as,
\begin{equation}\label{equ:linear_interpolation}
{\mathbf{f}_{t+1}} = (1-\eta){\mathbf{f}_{t}} + \eta{\mathbf{f}^{*}}
\end{equation}
where $\eta$ denotes the on-line adaptation rate.

Based on the formulation and optimization of BACF \cite{BACF}, we further explain its motivations to the extension of CFNet~\cite{CFNet} and the joint off-line learning of deep representation and model adaptation:
\begin{enumerate}
\item In CFNet, the deep representation is integrated with the simplest CF tracker~\cite{henriques2015high} for offline training.
    Note that many advanced CF models, e.g., BACF \cite{BACF}, can significantly outperform the simple CF in terms of tracking accuracy.
    Thus, it is natural to conjecture that the combination of deep representation and BACF can result in improved tracking performance.
\item One reason that CFNet only considers the conventional CF is that it has closed-form solution and can be interpreted as a differentiable CNN layer.
    As for BACF, the solution to Eqn.~(\ref{eq:bacf}) defines an implicit function of the feature representation $\mathbf{z}_t$ and model parameter $\lambda$, restricting its integration with CNN representation.
    Fortunately, when the number of iterations is fixed (i.e., truncated inference \cite{zuo2015discriminative,yan2016deep}), the $\mathbf{f}_{t+1}$ from Eqns.~(\ref{eq:admm}) and (\ref{equ:linear_interpolation}) can then be represented as an explicit function of the feature representation and model parameter.
    Therefore, by unrolling the ADMM optimization of BACF, it is feasible to facilitate the end-to-end off-line learning of truncated inference for visual tracking.
\item Moreover, BACF is performed on the handcrafted features in~\cite{BACF}.
    Denote by $\psi(\,\cdot \,; \mathbf{W}_F)$, a fully convolutional network with parameters $\mathbf{W}_F$.
    Thus, by letting $\mathbf{z}_t = \psi(\mathbf{x}_t; \mathbf{W}_F)$, both deep representation and truncated inference can be jointly off-line learned from annotated sequences.
\end{enumerate}
Motivated by the above discussions, we in the following first introduce a bi-level optimization framework for joint learning of deep representation and truncated inference, and then present the architecture and learning of our RTINet.
%
%
%--------------------------------------------------------------------------
\subsection{Model Formulation}
\label{sec:model_formulation}
Suppose $\mathbf{z}_t = \psi(\mathbf{x}_t; \mathbf{W}_F)$ is the deep representation of $\mathbf{x}_t$, where $\mathbf{W}_F$ denotes the parameters of the representor network $\psi(\, \cdot \,; \mathbf{W}_F)$.
Naturally, we require that the learned CFs $\mathbf{f}_{t+1} = \eta \mathbf{f}^{*} + (1-\eta)\mathbf{f}_{t}$ should be effective in tracking the target of the future frame.
Thus, the integration of BACF and deep representation can be formulated as a bi-level optimization problem,
{%\color{red}
\begin{eqnarray} \label{eq:cnn_bacf}
\begin{split}
&\min_{\lambda,\rho,{\mathbf M},\eta} \left\| {\mathbf y}_{t+1} - \sum_{l=1}^{L} \mathbf{z}_{t+1,l} \star (\eta \mathbf{f}^{*}_{l} + (1-\eta)\mathbf{f}_{t,l} )\right\|^2,  \\
&\mbox{ s.t. }\, \mathbf{f}^{*} = \arg \min_{\mathbf{f}} \left\|\mathbf{y}_t - \sum_{l=1}^L\mathbf{z}_{t,l} \star \mathbf{f}_l \right\| ^2 + \lambda \|\mathbf{h}\|^2, \\
&\mbox{ s.t. }\, \mathbf{f}_l = \mathbf{M}^{\top} \mathbf{h}_l
\end{split}
\end{eqnarray}}
However, $\mathbf{f}^{*}$ defines an implicit function of $\mathbf{z}_t$, and $\mathbf{f}_{t+1}$, making it difficult to compute the gradient.

With the unrolled ADMM optimization, when the number of iterations $K$ is fixed, all the $\mathbf{f}^{(1)}$, $...$, $\mathbf{f}^{(K)}$, and $\mathbf{f}_{t+1}$ can be represented as the functions of $\mathbf{z}_t$, $\mathbf{y}_t$, and $\mathbf{f}_{t}$.
For joint learning of deep representation and truncated inference, we also slightly modify the BACF model and ADMM algorithm to make that the model parameters $\lambda$ and $\mathbf{M}$, algorithm parameters $\rho$ and $\eta$ are both iteration-wise and learnable, i.e., $\Theta = \{ \Theta^{(1)}, ..., \Theta^{(K)} \}$ with $\Theta^{(k)} = \{ \lambda^{(k)}, \mathbf{M}^{(k)}, \rho^{(k)}, \eta^{(k)} \}$.
To improve the robustness of the learned tracker, we require that $\mathbf{f}_{t+1}$ can also be applied to the $\left(t+1\right)$-th frame.
To ease the training, we further introduce $\mathbf{f}_{t+1}^{(k)} = \eta^{(k)} \mathbf{f}^{(k)} + (1 - \eta^{(k)}) \mathbf{f}_{t}$, and require that $\mathbf{f}_{t+1}^{(k)}$ also performs well.
Taking all the aforementioned factors into account, we present the whole RTINet model for joint learning of representation and truncated inference as
\begin{eqnarray} \label{eq:loss_RTI}
\min \mathcal{L}(\mathbf{W}_F, \Theta) = \sum_{k=1}^{K} \left\| {\mathbf y}_{t+1} - \sum_{l=1}^{L} \psi_l(\mathbf{x}_{t+1}; \mathbf{W}_F) \star \mathbf{f}_{t+1,l}^{(k)} \right\|^2
\end{eqnarray}
where
\begin{eqnarray} \label{eq:constraint2_RTI}
\mathbf{f}_{t+1}^{(k)} = F_{{Int}}(\mathbf{f}^{(k)}, \mathbf{f}_{t}; \eta^{(k)}) = \eta^{(k)} \mathbf{f}^{(k)} + (1 - \eta^{(k)}) \mathbf{f}_{t} \, ,
\end{eqnarray}
\begin{subequations} \label{eq:admm_solution}
\small
\begin{empheq}[left=\empheqlbrace]{align}
%  \begin{cases}
\label{eq:admm_solution_a}
{{\mathbf{h}}^{\left( k \right)}} & =
F_{\mathbf{h}}({\mathbf f}^{(k\!-\!1)}, {\mathbf g}^{(k\!-\!1)}; \lambda^{(k)}, \rho^{(k)}, {\mathbf M}^{(k)}) \\
\nonumber & = \left( \lambda^{(k)}{\mathbf I} \!+\! \rho^{(k)} \left( {\mathbf M}^{(k)}{{\mathbf M}^{(k)}}^\top \!\otimes\! {\mathbf I}_{L}\right)\right)^{\!-\!1} \!\!\rho^{(k)}\!\! \left( {\mathbf M}^{(k)} \!\otimes\! {\mathbf I}_{L} \right) \left( {\mathbf f}^{(k\!-\!1)} \!+\! {\mathbf g}^{(k\!-\!1)}\right) \\
\label{eq:admm_solution_b}
{{\mathbf{g}}_l^{\left( k \right)}} & =
F_{\mathbf{g}} ({{\mathbf{g}}^{\left( k -1 \right)}}, {{\mathbf{f}}^{\left( k -1\right)}}, {{\mathbf{h}}^{\left( k\right)}}; \mathbf{M}^{(k)}) \\
\nonumber & = {{\mathbf{g}}_l^{\left( k -1 \right)}} + {{\mathbf{f}}_l^{\left( k -1\right)}} - \mathbf{M}^{(k)\top} {{\mathbf{h}}_l^{\left( k\right)}} \\
\label{eq:admm_solution_c}
\hat{\mathbf{f}}^{(k)}_{l}  & =
F_{\mathbf{f}}({\mathbf{z}}_{t}, {\mathbf{y}}_{t}, {{\mathbf{g}}}^{(k)}, {{\mathbf{h}}}^{(k)}; \rho^{(k)},{\mathbf M}^{(k)})  \\
\nonumber & = \frac{{\hat{\mathbf{z}}_{t,l}}^{*} \circ {\hat{\mathbf q}}}{{\rho^{(k)} + \sum_{l=1}^{L}{\hat{\mathbf{z}}_{t,l}^{*}} \circ {\hat{\mathbf{z}}_{t,l}}}}, \ {\hat{\mathbf q}}= \rho^{(k)}{\hat{\mathbf{h}}}^{(k)}_{l}\!-\!\rho^{(k)}{\hat{\mathbf{g}}}^{(k)}_{l} + {\hat{\mathbf{z}}_{t,l}} \circ {\hat{\mathbf{y}}_{t}}
\end{empheq}
\end{subequations}
% \end{cases}
% \end{eqnarray}
where ${\hat{\cdot}}=\mathcal{F}(\cdot)$ denotes the FFT of a signal, $\otimes$ indicates the Kronecker product, ${\mathbf I}_{L}$ is an identity matrix of size $L \times L$ and ${\hat{\mathbf{h}}}^{(k)}_{l}= \mathcal{F}(\mathbf{M}^{(k)\top} {{\mathbf{h}}_l^{\left( k\right)}})$.
${\mathbf f}^{(k)}$ can be further obtained by the inverse FFT of ${\hat {\mathbf f}}^{(k)}$.
In the first iteration, $\mathbf{f}^{(0)}$ and $\mathbf{g}^{(0)}$ are initialized as zeros.

To sum up, our RTINet consists of two subnetworks: (i) a representor network to generate deep representation $\mathbf{z}_t = \psi(\mathbf{x}_t; \mathbf{W}_F)$, and (ii) an updater network to update the CF model $\mathbf{f}_{t+1} = \mathbf{f}_{t+1}^{(K)} = \phi(\mathbf{z}_t, \mathbf{y}_t, \mathbf{f}_t; \Theta)$.
While the representor network  adopts the architecture of fully convolutional network, the updater network is recursively defined based on Eqns.~(\ref{eq:constraint2_RTI})$\sim$(\ref{eq:admm_solution_c}).
More detailed explanation on the representor and updater architecture will be given in the next subsection.
\begin{figure}[t]
\setlength{\abovecaptionskip}{-8pt}
\setlength{\belowcaptionskip}{-8pt}
\begin{center}
\includegraphics[width=0.85\linewidth]{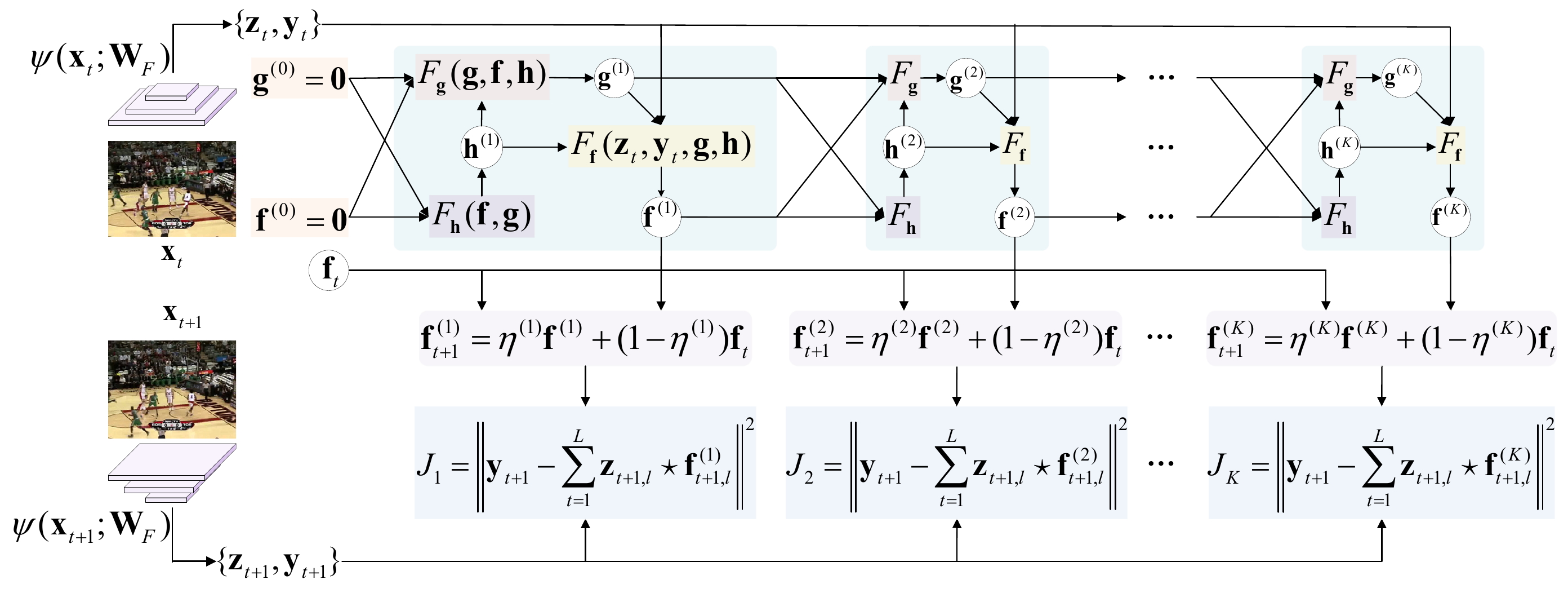}
\end{center}
\caption{Overview of the RTINet architecture, which includes a representor network and an updater network.
In the inference learning, we compute ${\mathbf h}$, ${\mathbf g}$ and ${\mathbf f}$ recursively following Eqns.~(\ref{eq:constraint2_RTI})$\sim$(\ref{eq:admm_solution_c}) in each stage.}
\label{fig:architecture}
\end{figure}
%
%
%---------------------------------------------------------------------
\subsection{Architecture of RTINet}
\label{sec:architecture}
Fig.~\ref{fig:architecture} provides an overview of the RTINet architecture,
which includes a representor network and a updater network.
For the representor network $\psi(\cdot;{\mathbf W}_{F})$, we adopt the first three convolution (conv) layers of the VGG-M \cite{VGGM}.
%
%We use the first three conv layers of VGG-M and update them further during joint learning with the updater network.
%
ReLU non-linearity and local response normalization are employed after each convolution operation,
and the pooling operation is deployed for the first two conv layers.
To handle different sizes of targets, we resize the patches to $224\times224$ as inputs and
produce the feature map with the size of {$13 \times 13 \times 512$}.

As for the updater network $\phi(\mathbf{z}_t, \mathbf{y}_t, \mathbf{f}_t; \Theta)$, we follow the unrolled ADMM optimization to design the network architecture.
%
% XH change:
As shown in Fig.~\ref{fig:architecture}, given $\{ \mathbf{z}_t, \mathbf{y}_t\}$, we initialize $\mathbf{f}^{(0)} = \mathbf{0}$ and $\mathbf{g}^{(0)} = \mathbf{0}$.
In the first stage of the updater network, (i)  the node $F_{\mathbf{h}} ({\mathbf{f}}, {\mathbf{g}})$ takes ${\mathbf{f}}^{(0)}$ and ${\mathbf{g}}^{(0)}$ as input to generate ${\mathbf{h}}^{(1)}$,
(ii) the node $F_{\mathbf{g}} ({\mathbf{g}}, {\mathbf{f}}, {\mathbf{h}})$ takes ${\mathbf{g}}^{(0)}$, ${\mathbf{f}}^{(0)}$, and ${\mathbf{h}}^{(1)}$ as input to generate ${\mathbf{g}}^{(1)}$,
and finally (iii) the node $F_{\mathbf{f}} ({\mathbf{z}}, {\mathbf{y}}, {\mathbf{g}}, {\mathbf{h}})$ takes ${\mathbf{z}}_t$, ${\mathbf{y}}_t$, ${\mathbf{g}}^{(1)}$ and ${\mathbf{h}}^{(1)}$ as input to generate ${\mathbf{f}}^{(1)}$.
By repeating $K$ stages, we can obtain ${\mathbf{f}}^{(K)}$, and then the node $F_{Int}(\mathbf{f}, \mathbf{f}_{t})$ takes ${\mathbf{f}}^{(K)}$ and $\mathbf{f}_{t}$ as input to generate $\mathbf{f}_{t+1}$.
Note that all the nodes $F_{\mathbf{g}}$, $F_{\mathbf{h}}$, $F_{\mathbf{f}}$, and $F_{Int}$ are differentiable.
Thus, with the annotated video sequences, both the updater network and the representor network can be end-to-end trained by minimizing the model objective in Eqn.~(\ref{eq:loss_RTI}).
%----------------------------------------------------------------------------
\subsection{Model Learning}
\label{sec:learning}
In this subsection, we present a stage-wise learning scheme to learn the model parameters ${\mathbf W}_{F}$ and $\Theta = \{{\Theta}^{(k)}\}_{k=1,2,\cdots,K}$.
After the first $(k^{\prime}-1)$ stages of learning, we can obtain the current model parameters ${\mathbf W}_{F}$ and $\{ {\Theta}^{(k)}\}_{k=1,2,\cdots,{(k^{\prime}-1)}}$.
Denote by $\Theta^{(k^{\prime})}=\{\lambda^{(k^{\prime})},{\mathbf M}^{(k^{\prime})},\rho^{(k^{\prime})},\eta^{(k^{\prime})}\}$.
To guide the model learning, we define the stage-wise loss function as,
\begin{eqnarray} \label{eq:loss_stage_k}
J_{k^{\prime}} = \left\| {\mathbf y}_{t+1} - \sum_{l=1}^{L} \mathbf{z}_{t+1,l} \star {\mathbf f}^{(k^{\prime})}_{t+1,l} \right\|^2 \, .
\end{eqnarray}
Then we introduce the gradient computation which is used to update model parameters with the stochastic gradient descent (SGD) algorithm.

According to Eqns.~(\ref{eq:constraint2_RTI})$\sim$(\ref{eq:admm_solution_c}), we have the following observations:
\begin{enumerate}[(a)]
\item ${\mathbf f}^{\left(k^{\prime} \right)}_{t+1}$ is a function of ${\mathbf f}^{(k^{\prime})}$, ${\mathbf f}_{t}$ and $\eta^{\left(k^{\prime}\right)}$;
\item ${\mathbf h}^{\left(k^{\prime}\right)}$ is a function of ${\mathbf f}^{\left(k^{\prime}-1\right)}$, ${\mathbf g}^{\left(k^{\prime}-1\right)}$, $\lambda^{\left(k^{\prime}\right)}$, $\rho^{\left(k^{\prime}\right)}$ and ${\mathbf M}^{(k^{\prime})}$;
\item ${\mathbf g}^{\left(k^{\prime}\right)}$ is a function of ${\mathbf g}^{\left(k^{\prime}-1\right)}$, ${\mathbf f}^{\left(k^{\prime}-1\right)}$, ${\mathbf h}^{\left(k^{\prime}\right)}$ and ${\mathbf M}^{\left(k^{\prime}\right)}$;
\item ${\mathbf f}^{\left(k^{\prime}\right)}$ is a function of ${\mathbf z}_{t}$, ${\mathbf y}_{t}$, ${\mathbf h}^{\left(k^{\prime}\right)}$, ${\mathbf g}^{\left(k^{\prime}\right)}$, $\rho^{\left(k^{\prime}\right)}$ and ${\mathbf M}^{\left(k^{\prime}\right)}$.
\end{enumerate}
Combined these observations with Eqn.~(\ref{eq:loss_stage_k}), we can obtain the gradient of $J_{k^{\prime}}$ w.r.t. $\Theta^{(k^{\prime})}$ in the $k^{\prime}$-th stage, i.e.,
%\begin{eqnarray} \label{eq:gd_theta}
$\nabla_{\!\Theta^{\left(\!k^{\prime}\!\right)}}J_{k^{\prime}} \!\!=\!\!
\left( \!
\nabla_{\!\eta^{\left(\!k^{\prime}\!\right)}}J_{k^{\prime}}, \! \!
\nabla_{\!\rho^{\left(\!k^{\prime}\!\right)}}J_{k^{\prime}}, \! \!
\nabla_{\!{\mathbf M}^{\left(\!k^{\prime}\!\right)}}J_{k^{\prime}}, \! \!
\nabla_{\!\lambda^{\left(\!k^{\prime}\!\right)}}J_{k^{\prime}}
\!\right).$
%\end{eqnarray}
%

Specifically, for each parameter in $\Theta^{\left(k^{\prime}\right)}$, we have,
\begin{eqnarray} \label{eq:gd}
 \begin{cases}
% \eta
\nabla_{\eta^{\left(k^{\prime}\right)}} J_{k^{\prime}} & \! \! \! \! \! \! \! = \! \!
\nabla_{{\mathbf f}^{\left(k^{\prime}\right)}_{t+1}} J_{k^{\prime}}
\nabla_{\eta^{\left(k^{\prime}\right)}} {\mathbf f}^{\left(k'\right)}_{t+1} \\
% \rho
\nabla_{\rho^{\left(k^{\prime}\right)}} J_{k^{\prime}} & \! \! \! \! \! \! \! = \! \!
\nabla_{{\mathbf f}^{\left(k^{\prime}\right)}} J_{k^{\prime}}
\nabla_{\rho^{\left(k^{\prime}\right)}} {\mathbf f}^{\left(k^{\prime}\right)} +
\nabla_{{\mathbf h}^{\left(k^{\prime}\right)}} J_{k^{\prime}}
\nabla_{\rho^{\left(k^{\prime}\right)}} {\mathbf h}^{\left(k^{\prime}\right)}
\\
% M
\nabla_{{\mathbf M}^{\!\left(k^{\prime}\!\right)}} J_{k^{\prime}} & \! \! \! \! \! \! \! = \! \!
\nabla_{{\mathbf f}^{\!\left(k^{\prime}\!\right)}} J_{k^{\prime}}
\nabla_{{\mathbf M}^{\!\left(k^{\prime}\!\right)}} {\mathbf f}^{\!\left(k^{\prime}\!\right)} \!\!+\!\!
\nabla_{{\mathbf g}^{\!\left(k^{\prime}\!\right)}} J_{k^{\prime}}
\nabla_{{\mathbf M}^{\!\left(k^{\prime}\!\right)}} {\mathbf g}^{\!\left(k^{\prime}\!\right)} \!\!+\!\!
\nabla_{{\mathbf h}^{\!\left(k^{\prime}\!\right)}} J_{k^{\prime}}
\nabla_{{\mathbf M}^{\!\left(k^{\prime}\!\right)}} {\mathbf h}^{\!\left(k^{\prime}\!\right)} \\
% \lambda
\nabla_{\lambda^{\left(k^{\prime}\right)}} J_{k^{\prime}} & \! \! \! \! \! \! \! = \! \!
\nabla_{{\mathbf h}^{\left(k^{\prime}\right)}_{t+1}} J_{k^{\prime}} \cdot
\nabla_{\lambda^{\left(k^{\prime}\right)}} {\mathbf h}^{\left(k^{\prime}\right)}_{t+1}
\end{cases}
\end{eqnarray}
The derivations of $\nabla_{{\mathbf f}^{\left(k^{\prime}\right)}} J_{k^{\prime}}$, $\nabla_{{\mathbf g}^{\left(k^{\prime}\right)}} J_{k^{\prime}}$ and $\nabla_{{\mathbf h}^{\left(k^{\prime}\right)}} J_{k^{\prime}}$ are presented in the supplementary materials.

Furthermore, $J_{k^{\prime}}$ should also be used to update the model parameters ${\mathbf W}_{F}$ and $\{ {\Theta}^{(k)}\}_{k=1,2,\cdots,{(k^{\prime}-1)}}$ for the sake of joint representation and truncated inference learning.
Thus, we also give the gradient of $J_{k^{\prime}}$ w.r.t. $\mathbf{h}^{\left(k^{\prime}-1\right)}$, $\mathbf{g}^{\left(k^{\prime}-1\right)}$, and
$\mathbf{f}^{\left(k^{\prime}-1\right)}$ as follows,
\begin{eqnarray} \label{eq:gd_2}
 \begin{cases}
 % h
\nabla_{{\mathbf h}^{\left(k^{\prime}-1\right)}} J_{k^{\prime}} =
\nabla_{{\mathbf g}^{\left(k^{\prime}-1\right)}} J_{k^{\prime}}
\nabla_{{\mathbf h}^{\left(k^{\prime}-1\right)}} {\mathbf g}^{\left(k^{\prime}-1\right)} +
\nabla_{{\mathbf f}^{\left(k^{\prime}-1\right)}} J_{k^{\prime}}
\nabla_{{\mathbf g}^{\left(k^{\prime}-1\right)}} {\mathbf f}^{\left(k^{\prime}-1\right)} \\
% g
\nabla_{{\mathbf g}^{\left(k^{\prime}-1\right)}} J_{k^{\prime}} =
\nabla_{{\mathbf g}^{\left(k^{\prime}\right)}} J_{k^{\prime}}
\nabla_{{\mathbf g}^{\left(k^{\prime}-1\right)}} {\mathbf g}^{\left(k^{\prime}\right)} +
\nabla_{{\mathbf h}^{\left(k^{\prime}\right)}} J_{k^{\prime}}
\nabla_{{\mathbf g}^{\left(k^{\prime}-1\right)}} {\mathbf h}^{\left(k^{\prime}\right)} \\
% f
\nabla_{{\mathbf f}^{\left(k^{\prime}-1\right)}} J_{k^{\prime}} =
\nabla_{{\mathbf g}^{\left(k^{\prime}\right)}} J_{k^{\prime}}
\nabla_{{\mathbf f}^{\left(k^{\prime}-1\right)}} {\mathbf g}^{\left(k^{\prime}\right)} +
\nabla_{{\mathbf h}^{\left(k^{\prime}\right)}} J_{k^{\prime}}
\nabla_{{\mathbf f}^{\left(k^{\prime}-1\right)}} {\mathbf h}^{\left(k^{\prime}\right)}
\end{cases}
\end{eqnarray}
Please refer to the supplementary material for the detail of the derivation.
Therefore, we can back-propagate the gradient to the $(k^{\prime}-1), ..., 1$ layers and the representor network $\psi(\cdot; \mathbf{W}_F)$.
After the learning of the $k^{\prime}$-th stage, we can further conduct the $(k^{\prime}+1)$-th stage-wise training by learning $\Theta^{(k^{\prime}+1)}$ and fine-tuning ${\mathbf W}_{F}$ and $\{ {\Theta}^{(k)}\}_{k=1,2,\cdots,{k^{\prime}}}$ until the ending of the $K$-th stage-wise training.
Finally, all the model parameters ${\mathbf W}_{F}$ and $\Theta$ are adopted for target localization and model adaptation during the on-line tracking process.

%
 \iffalse
%---------------------------------------------------------------------------------
\subsection{Online Tracking with RTINet}
\label{sec:tracking}
To apply our RTINet to online tracking, we first use the representor network $\psi(\cdot; \mathbf{W}_F)$
to extract feature representation of the first frame, and utilize the annotated target bounding box
and the BACF model~\cite{BACF} to initialize the CF $\mathbf{f}_1$.
%
When the $t$-th ($t > 0$) frame comes, after extracting its representation using $\psi(\cdot; \mathbf{W}_F)$,
the current CF $\mathbf{f}_t$ is first deployed to localize the target,
and the updater network $\phi(\mathbf{z}_t, \mathbf{y}_t, \mathbf{f}_t; \Theta)$ is then used to update $\mathbf{f}_{t+1}$.
%

To penalize large displacements, we multiply a cosine window with the response map during tracking.
%
Moreover, the object is searched over five scales $1.01^{\{-2, -1, 0, 1,2\}}$ to handle the scale variations,
while the specific scale with the highest value is considered as the tracking result in current frame.
%
\fi
%--------------------------------------------------------------------------------------
%--------------------------------------------------------------------------------------
\section{Experiments}
\label{sec:experiments}
In this section, we first describe the implementation details,
%
%then compare with the baseline trackers those are highly relevant to
%our approach on the OTB-2015 dataset, and its subset TB-50, which consists of 50 challenging sequences.
{then compare with the baseline trackers highly relevant to our approach.}
For comprehensive analysis, ablation studies are conducted to investigate
the effect of the joint feature representation learning and stage-wise training scheme.
Finally, we compare the proposed RTINet with state-of-the-art trackers
on the OTB-2015 \cite{wu2015object}, TB-50 \cite{wu2015object} {(i.e., the 50 more challenging sequences from OTB-2015)},
TempleColor-128 \cite{liang2015encoding}
and VOT2016 \cite{Kristan2016a} datasets. % to demonstrate the effectiveness of the proposed RTINet tracker.
Our approach is implemented in MATLAB 2017a using MatConvNet library, and all the experiments are run
on a PC equipped with an Intel i7 CPU 4.0GHz, 32GB and a single NVIDIA GTX 1080 GPU.
%
%The code and results will be publicly available.
%

%
%-------------------------------------------------------------------------------
\subsection{Implementation Details}
\label{sec:implementation_details}
{\flushleft {\bf Training Set.}}
To train the RTINet, we employ the 2015 edition of ImageNet Large Scale Visual Recognition Challenge (ILSVRC2015) dataset,
which consists of more than 4,500 videos from 30 different object categories.
For each video, we pick up 20 successive frames in which the target sizes are not larger than 50\% of the image size.
Then, 2,000 sequences are randomly chosen for training and the rest are used as the validation set.
To avoid the influence of target distortion, we crop the square region centered at the target
with the size of $5\sqrt{WH} \times 5\sqrt{WH}$, where $W$ and $H$ represent the width and height of the target, respectively.
And the cropped regions are further resized to $224\times 224$ as the input of the RTINet.
{\flushleft {\bf Training Details.}}
Since it is not trivial to train the RTINet with all the parameters directly,
we decouple the training of the representor network and updater network into two steps:
(1) We firstly keep the representor network fixed and train the updater network in a greedily stage-wise manner.
As for the stage $k$, we initialize the hyper-parameters of the updater network
(i.e., $\lambda^{(k)}$, $\rho^{(k)}$, $\eta^{(k)}$ and $\mathbf{M}^{(k)}$) with the trained parameters in the previous stage $k-1$.
Then the updater network is trained with 50 epochs with all the parameters in the previous stages fixed.
(2) After the stage-wise training of the updater network, we apply another 50 epochs to jointly train the representor network and updater network.

During training, we initialize the convolution layers of the representor network with the pre-trained VGG-M model~\cite{VGGM}.
As for the model parameters, we set $\lambda^{(0)}$, $\rho^{(0)}$, $\eta^{(0)}$ and $\mathbf{M}^{(0)}$
in the first stage of the updater network as {1, 1, 0.013} and the binary selection matrix, respectively.
We use the stochastic gradient descent (SGD) as the optimizer with the mini-batch size of 16,
and the learning rate is exponentially decayed from $10^{-2}$ to $10^{-5}$.
%
%--------------------------------------------------------------------------------
%
%%%%%%%%%%%%%%%%%%%%%%%%%%%%%%%%%%%%%%%%%%%%%%%%%%%%%%%%%%%%%%%%%%%%%%
%
\begin{table*}[htbp]
\caption{Comparison with the baseline CFNet variants on OTB-2015.}
\label{tab:baseline_cfnet}
\centering
\resizebox{\textwidth}{!}{
\begin{tabular}{lcccccc}
\toprule
Trackers &
{CFNet-conv1} &
{CFNet} &
{CFNet-conv1-Rep} &
{CFNet-Rep} &
{RTINet-conv1} &
{RTINet}\\
\midrule
AUC   & 53.6 & 56.8 & 54.8 & 58.0 & {\color{blue}64.3} & {\color{red}68.2} \\
FPS   & {\color{red}84} & {75} & {\color{blue}82.7} & 68 & 23.3 &     9.0\\
\bottomrule
\end{tabular}
}
\end{table*}
\begin{table*}[htbp]
\caption{Comparison with the baseline BACF variants on OTB-2015.}
\label{tab:baseline_bacf}
\centering
\resizebox{\textwidth}{!}{
\begin{tabular}{lccccccc}
\toprule
Trackers   &
{BACF} &
{BACF-VGGM} &
{BACF-Rep}  &
{RTINet-VGGM} &
{stdBACF-Rep} &
{RTINet} \\
\midrule
AUC   & {61.5} & 63.1 & {64.0}  & {\color{blue}66.5} & 64.2 & {\color{red}68.2}  \\
FPS   & 35.3     & 6.1 & 6.5 & {\color{blue}8.9} & {7.0} & {\color{red}9.0}  \\
\bottomrule
\end{tabular}
}
\end{table*}
\subsection{Comparison with CFNet}
The most relevant methods to our RTINet is CFNet~\cite{CFNet},
which is also proposed for the joint learning of deep representation and CF tracker.
In comparison, the updater network of our RTINet is designed based on the unrolled optimization of BACF~\cite{BACF}.
Here, we evaluate two variants of the proposed method: RTINet with three convolution layers and its rapid version,
i.e., RTINet-conv1 with one convolution layer,
{and compare them with CFNet, CFNet-conv1,
and their two variants with features extracted by RTINet representor,
i.e., CFNet-conv1-Rep and CFNet-Rep on OTB-2015.}
{Following the protocols in \cite{wu2015object}, we report the results in terms of area under curve (AUC)
and tracking speed in Table~\ref{tab:baseline_cfnet}.}
And we have {two} observations.
{
(1) The CFNet variants with RTINet features perform better than CFNet-conv1 and CFNet with an AUC gain of 1.2\% and 1.2\%, respectively, thereby showing the effectiveness and generalization of the deep features learned by RTINet.}
(2) In terms of {AUC}, both RTINet variants perform favorably against their counterparts,
indicating that RTINet is effective in learning feature representation and truncated inference.
In particular, RTINet brings an AUC gain of {11.4}\% over CFNet on the OTB-2015 dataset.
As for the rapid version, RTINet-conv1 also outperforms its baseline CFNet-conv1 by a gain of {10.7}\%.
%
%{\color{red}It can be noted that both RTINet-conv1 and RTINet show extreme superiority to the CFNet variants.}
%
RTINet even achieves an AUC of 68.2\% on OTB-2015, outperforming other trackers with a large margin.
We owe the improvements to both the introduction of the advanced BACF tracker and truncated inference into the RTINet framework.
%
%(3) The proposed RTINet is superior to the BACF tracker which uses HOG feature,
%which can be explained by the discriminative ability of the learned representation and truncated inference.
%

We also report the average FPS of different trackers.
While the best speed belongs to the CFNet-conv1 (84 fps) and CFNet-conv1-Rep (82.7 fps),
%
%the tracking accuracy of these two methods are much lower than ours.
%
RTINet runs at 9 fps and achieves the state-of-the-art tracking accuracy.
Actually, a large part of computational cost in RTINet comes from the deeper CNN feature extraction.
When conv1 feature is adopted, and RTINet-conv1 achieves a real time speed of 24 fps
while still performing favorably against CFNet.
%

%
%%%%%%%%%%%%%%%%%%%%%%%%%%%%%%%%%%%%%%%%%%%%%%%%%%%%%%%%%%%%%%%%%%%%%%
%
\begin{table*}[htbp]
\caption{The AUC scores of RTINet by training with different number of stages.}
\label{tab:stage_training}
\centering
%\resizebox{0.8\textwidth}{!}{
\begin{tabular}{p{3cm}p{0.8cm}p{0.8cm}p{0.8cm}p{0.8cm}p{0.8cm}p{0.8cm}p{0.8cm}p{0.8cm}p{0.8cm}p{0.8cm}}
\toprule
Number of Stages   &1 &2 &3 &4 &5 &6 &7  &8 &9 &10 \\
\midrule
Basketball   & 62.0   &{75.9} & {69.1} & 64.3& {69.4} &{69.1} &68.9 &68.9 &68.8 &68.8 \\
BlurCar1   &77.1  & {83.0} & {81.2} &{81.1} &80.6 &80.7 & 80.5& 80.4& 80.3&80.3 \\
CarDark   &76.2  & {85.7} &{83.3} & {82.9}& 82.2& 82.1&81.6 &81.7 & 82.2& 82.3\\
Human4   &44.1  & {57.0} & 55.6 &{57.7} & {61.5}& 51.0 &52.2 & 51.5& 52.0& 52.3\\
%Panda   &48.3  &50.7 & 48.8& 52.0& \textbf{52.2}& \textbf{52.2} & 52.1& 47.8& 51.8& 51.8\\
%Suv   &49.3  &77.9 & 77.8&77.8 &77.9 &78.0 & 78.1&\textbf{78.2} &\textbf{ 78.2} &78.0 \\
Toy   &60.1  & 61.1 &{63.1}& {62.8}& 62.1& 61.9& {62.8}& {62.8}& 62.7& {63.0} \\
%Walking   &  & 73.4& 73.6& 73.8& 73.9& 73.8& 74.3& 74.3& 74.0& \textbf{74.5}\\
\midrule
OTB-2015    & 59.6  & {68.2} & {67.2}& {67.2} & {66.3} &66.0 &65.6 & 66.3& 66.0& 66.2 \\
\bottomrule
\end{tabular}
%}
\end{table*}
%
%%%%%%%%%%%%%%%%%%%%%%%%%%%%%%%%%%%%%%%%%%%%%%%%%%%%%%%%%%%%%%%%%%%%%%
%-------------------------------------------------------------------------------
\subsection{Ablation studies}
\label{sec:ablative_study}
In this section, we analyze in depth the effect of joint feature representation and truncated inference learning as well as stage-wise training.
%
%
%-------------------------------
{\flushleft {\bf Joint Learning.}}
To investigate the effect of joint learning, we decouple the feature representation and truncated inference learning,
which results in four variants of RTINet:
BACF-VGGM (BACF with the fixed convolutional feature from pre-trained VGG-M),
BACF-Rep (BACF with the learned RTINet representation),
RTINet-VGGM (RTINet with the fixed convolution feature from pre-trained VGG-M)
and the full RTINet model.
{Besides, we also apply the learned RTINet representation and model parameters $\lambda$, $\eta$ and ${\mathbf M}$ to the standard BACF, resulting in stdBACF-Rep.}
%
%Due to the different implementation with the standard BACF in above BACF based variants,
%we also compare with the standard BACF-Rep, namely stdBACF-Rep, which is applied with the learned $\lambda$, $\eta$ and ${\mathbf M}$.
%%
%
Table~\ref{tab:baseline_bacf} shows the AUC scores of the default BACF with HOG features, and the BACF variants on OTB-2015.
From Table~\ref{tab:baseline_bacf}, it can be seen that RTINet and RTINet-VGGM improve the AUC scores significantly in comparison with the corresponding BACF variants.
This can be attributed to that the truncated inference learning in updater network does benefit the tracking performance.
Moreover, RTINet also improves the performance of RTINet-VGGM by an AUC gain of {1.7}\%,
and BACF-Rep obtains a gain of {0.9}\% over BACF-VGGM, validating the
effectiveness of representation learning.
It is worth noting that, in our RTINet the inference learning improves
the performance more than the feature learning, implying that pre-trained VGG-M does have good representation and generalization ability.
To sum up, both the learned feature representation and truncated inference are helpful in improving tracking accuracy, which together explain the favorable performance of our RTINet.
%
%-------------------------------
{\flushleft {\bf Stage-wise Learning.}}
In Section~\ref{sec:method}, we present a stage-wise training scheme to learn model parameters.
In particular, we solve the BACF \cite{BACF} formulation using the truncated ADMM optimization.
Thus, we analyse the effect of stage number on tracking performance.
Table~\ref{tab:stage_training} gives the average AUC score of RTINet on all sequences as well as several representative ones by setting different number of stages on the OTB-2015 dataset.
{RTINet with one stage performs poorly
with the AUC of {59.6}\%, even lower than the BACF (61.5\%).}
This is reasonable due to that RTINet only with one stage is similar to the simple CF rather than the advanced BACF model.
{
Benefited from the advanced BACF, RTINet achieves significantly better
performance within 2$\sim$5 iterations for most sequences.
The best AUC score of {68.2}\% of RTINet is attained with two stages on OTB-2015, indicating that efficient solver can be learned.
It can also be found that increasing number of stages causes moderate decrease on AUC.
One possible reason is that for smaller number of stages, RTINet focuses on minimizing upper loss in Eqn.~(\ref{eq:cnn_bacf}) and benefits accuracy.
For larger number of stages, RTINet may begin to minimize lower loss in Eqn.~(\ref{eq:cnn_bacf}) instead of accuracy.}
\begin{figure}[htbp]
\centering
\subfigure[]{ \label{figure:1a}
\includegraphics[width=0.28\textwidth]{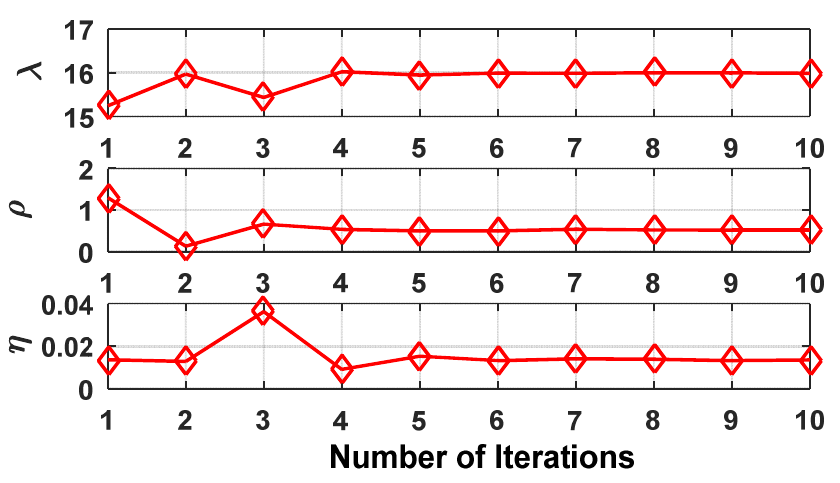}
}
\subfigure[]{ \label{figure:1b}
\includegraphics[width=0.35\textwidth]{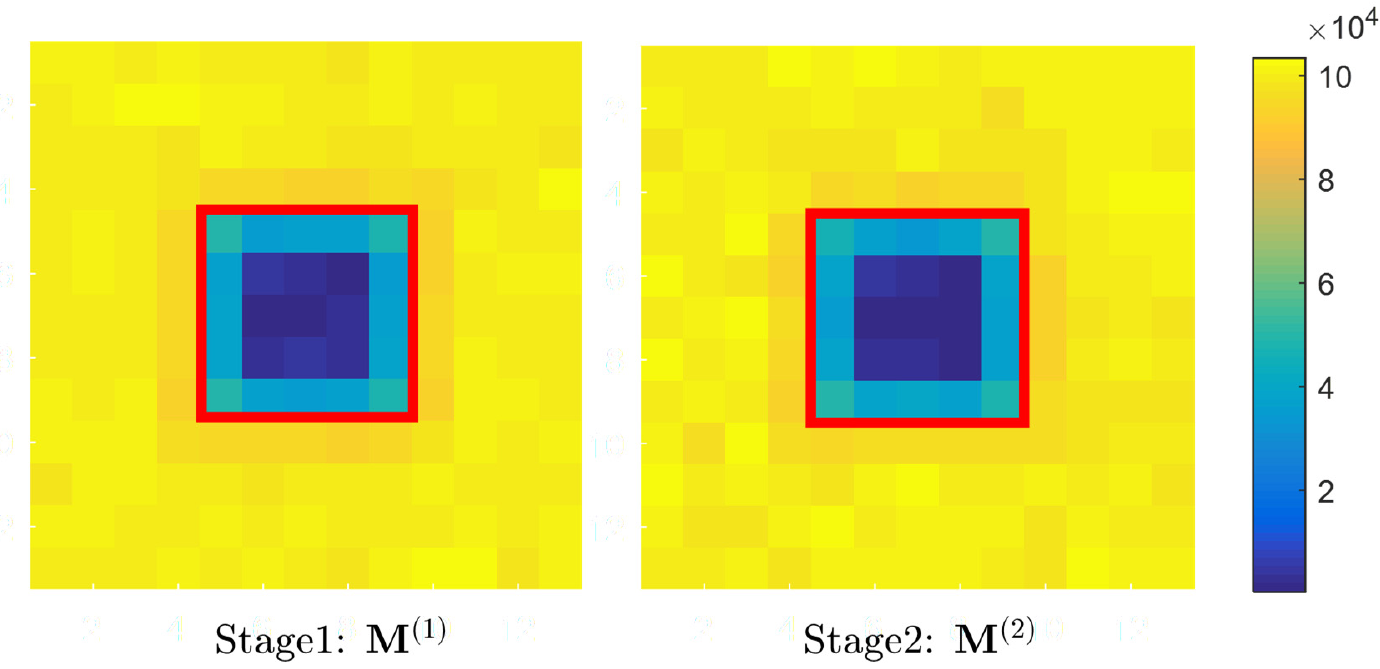}
}
\subfigure[]{ \label{figure:1a}
\includegraphics[width=0.29\textwidth]{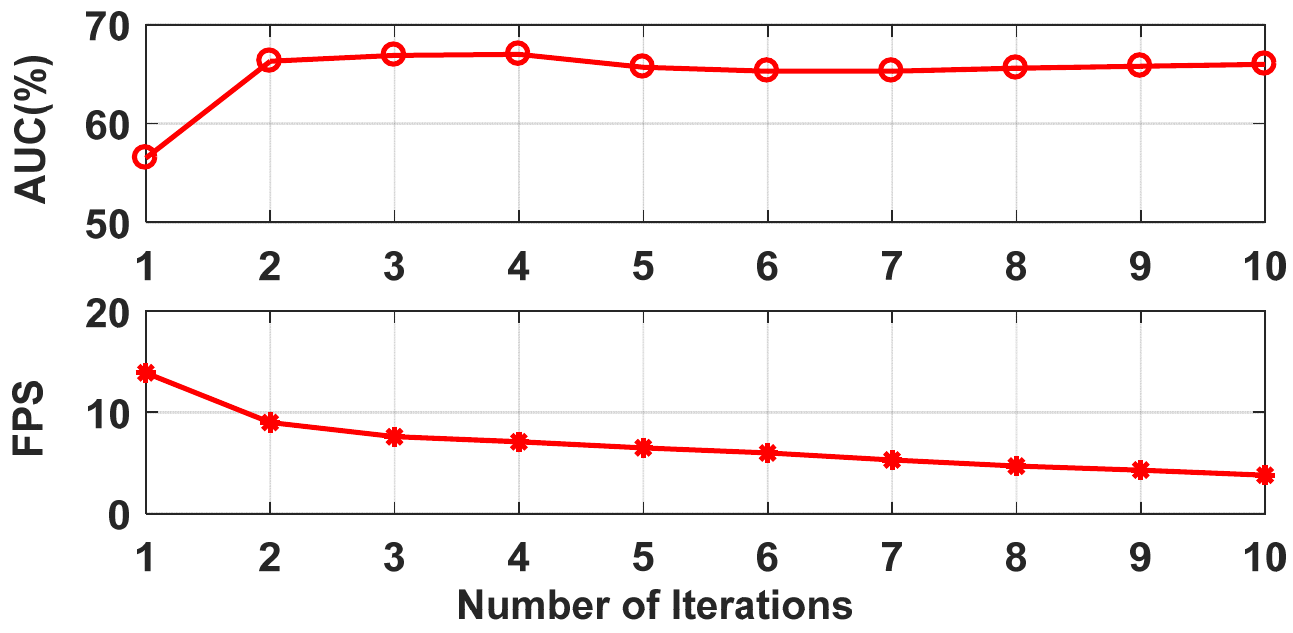}
}
\caption{\small{(a) The learned $\lambda,\rho,\eta$ for each stage. (b) Visualization of ${\mathbf M}$ for the first two stages. (c) Evaluation on the number of stages used for testing with an off-line trained 10-stage RTINet.}}
\label{fig:param_analysis}
\end{figure}
%
%%%%%%%%%%%%%%%%%%%%%%%%%%%%%%%%%%%%%%%%%%%%%%%%%%%%%%%%%%%%%%%%%%%%%%
%
%-------------------------------
{
{\flushleft {\bf Visualization of Learned Parameters.}}
Parameters at all stages are off-line trained and then keep fixed during tracking.
Fig.~\ref{fig:param_analysis}(a) shows the plots of the learned stage-wise $\lambda, \rho, \eta$
used in Table~\ref{tab:stage_training}.
It can be noted that the values of $\lambda, \rho, \eta$ become stable from the fourth stage.
From Table~\ref{tab:stage_training}, the best tracking accuracy is attained when the stage number is two.
Thus, we present the visualization of the learned ${\mathbf M}$s for the first two stages in Fig.~\ref{fig:param_analysis}(b).
%
%Corresponding to the two stages based RTINet which achieves the best tracking performance,
%the learned  parameter values are $\{\lambda^{(1)}\!=\!15.24500, \rho^{(1)}\!=\!1.29460,
%\eta^{(1)}\!=\!0.01363\}$, $\{\lambda^{(2)}\!=\!15.96700,  \rho^{(2)}\!=\!0.12740, \eta^{(2)}\!=\!0.01288\}$.
%The learned ${\mathbf M}$s of the two stages are visualized in Fig.~\ref{fig:param_analysis}(a).
%
From Fig.~\ref{fig:param_analysis}(a)(b), we have two observations:
(1) each stage has its specific parameter values,
(2) the learned ${\mathbf M}$s relax the binary cropping operation which is slightly different with the ${\mathbf M}$ adopted in BACF.
We also note that both the ${\mathbf M}$ in BACF and our learned ${\mathbf M}$s are resized to the feature map size in tracking.
%
%-------------------------------
{\flushleft {\bf Effects of Convergence on Tracking.}}
Generally, the ADMM algorithms are adopted to resolve the constrained convex optimization problem with a guarantee of convergence.
Thus, it is interesting to discuss the effect of iteration numbers after training RTINet with a fixed number of stages.
%
%Fig.~\ref{fig:convergence_analysis}(a) plots the duality gap w.r.t. number of iterations ($K$) on the second frame of the sequence {\textit{Matrix}}.
%
%The duality gap gradually decreases along with the increase of $K$, indicating that with larger $K$,
%RTINet can converge to better solution in terms of lower loss in Eqn.(\ref{eq:cnn_bacf}).
%
To this end, we train a 10-stage RTINet and test it on the OTB-2015 by using different number of iterations in tracking.
From Fig.~\ref{fig:param_analysis}(c), the best tracking accuracy is obtained after 4 iterations.
Then RTINet may focus on minimizing the lower loss and more iterations does not bring any increase on accuracy.
Fig.~\ref{fig:param_analysis}(c) also shows the plot of tracking speed.
Comparing Table~\ref{tab:stage_training} and Fig.~\ref{fig:param_analysis}(c), it can be seen that direct training RTINet with small $K$ is better than first training a 10-stage RTINet and then testing it with small iterations.
}
%
%%%%%%%%%%%%%%%%%%%%%%%%%%%%%%%%%%%%%%%%%%%%%%%%%%%%%%%%%%%%%%%%%%%%%%
\begin{figure*}[t] % htb
\centering
\subfigure[TB-50]{ \label{fig:otb2013_auc}
\includegraphics[width=0.3\textwidth]{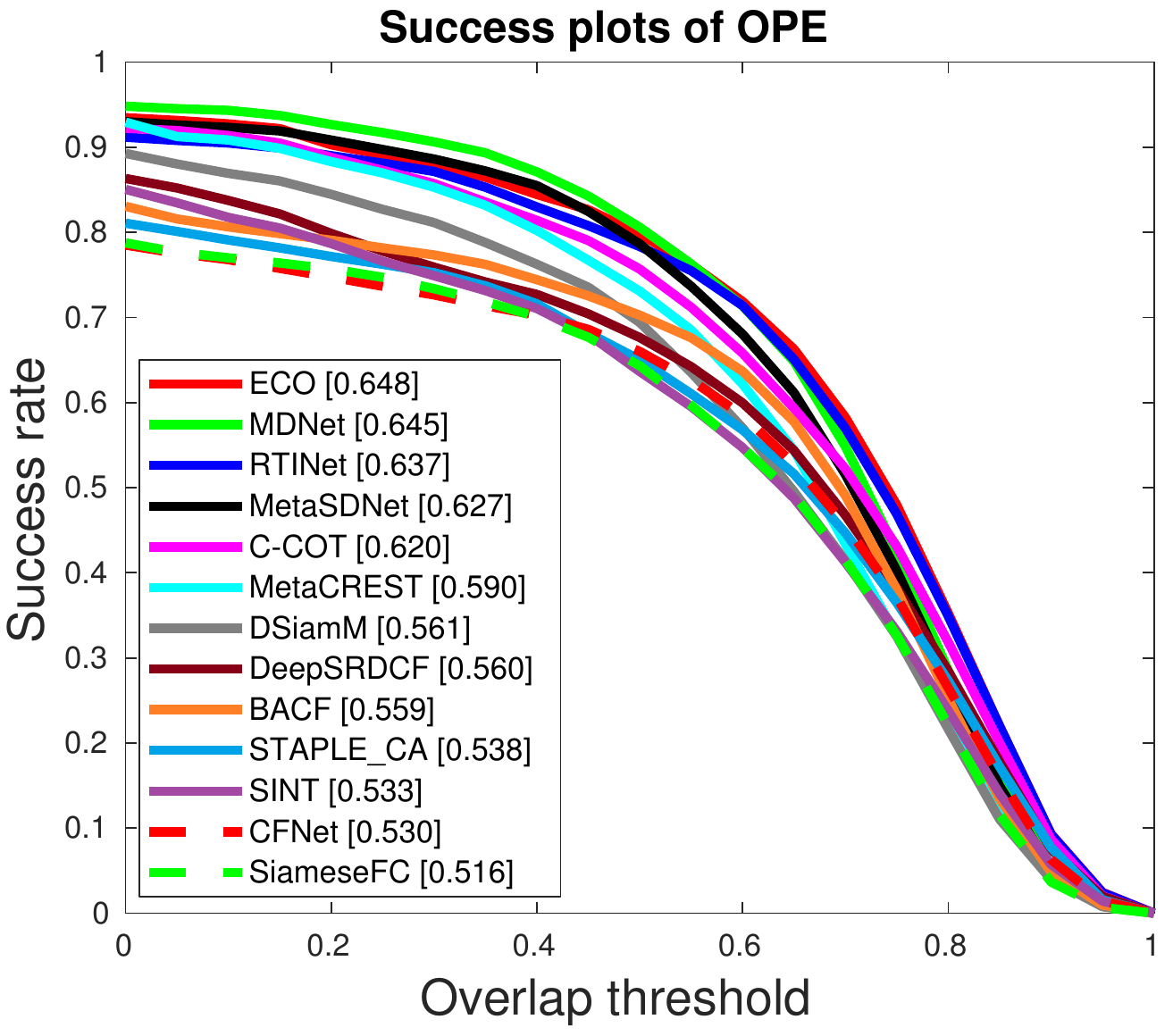}}
\subfigure[OTB-2015]{ \label{fig:otb2015_auc}
\includegraphics[width=0.3\textwidth]{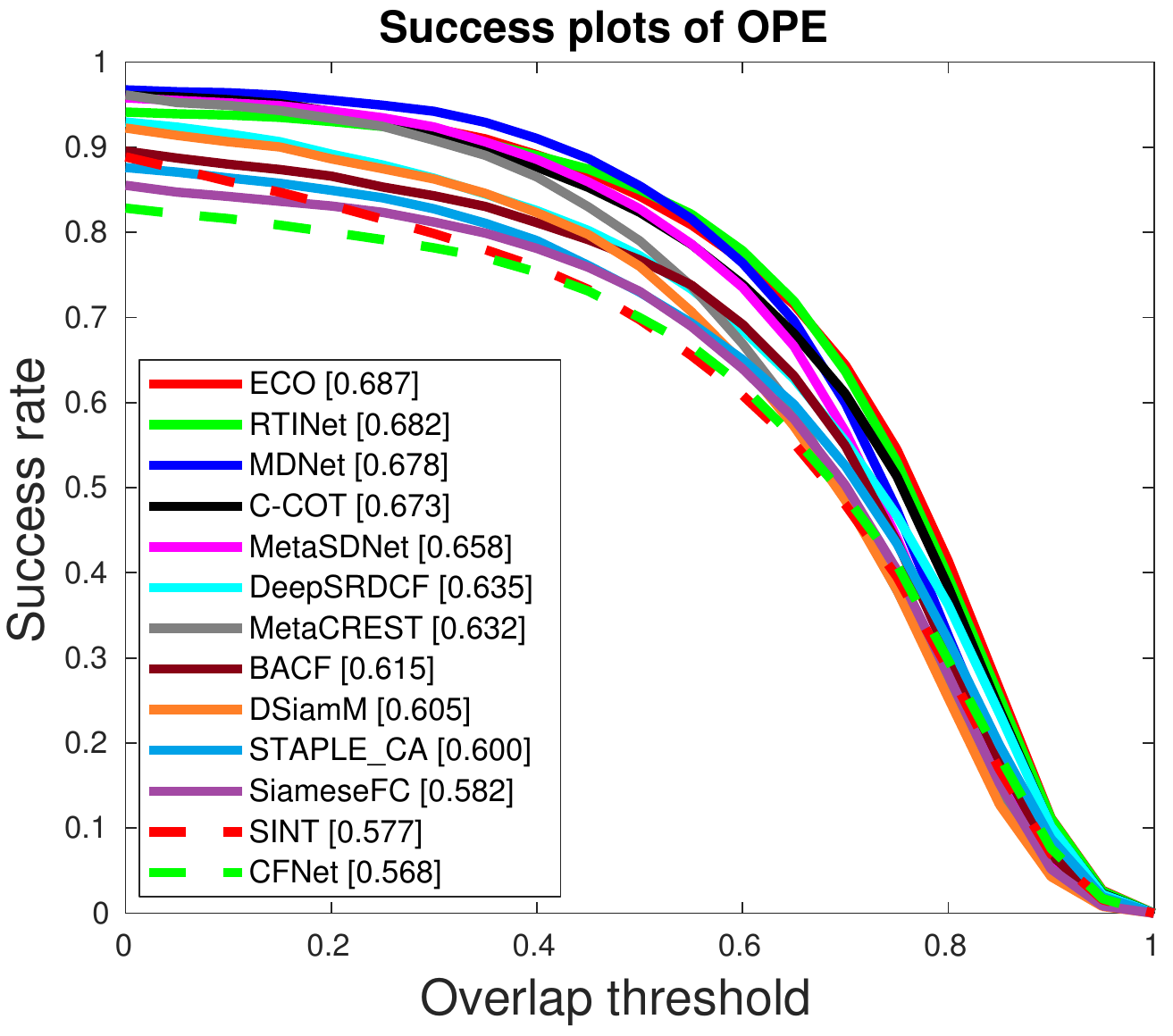}}
\subfigure[TempleColor-128]{ \label{fig:tc128_auc}
\includegraphics[width=0.3\textwidth]{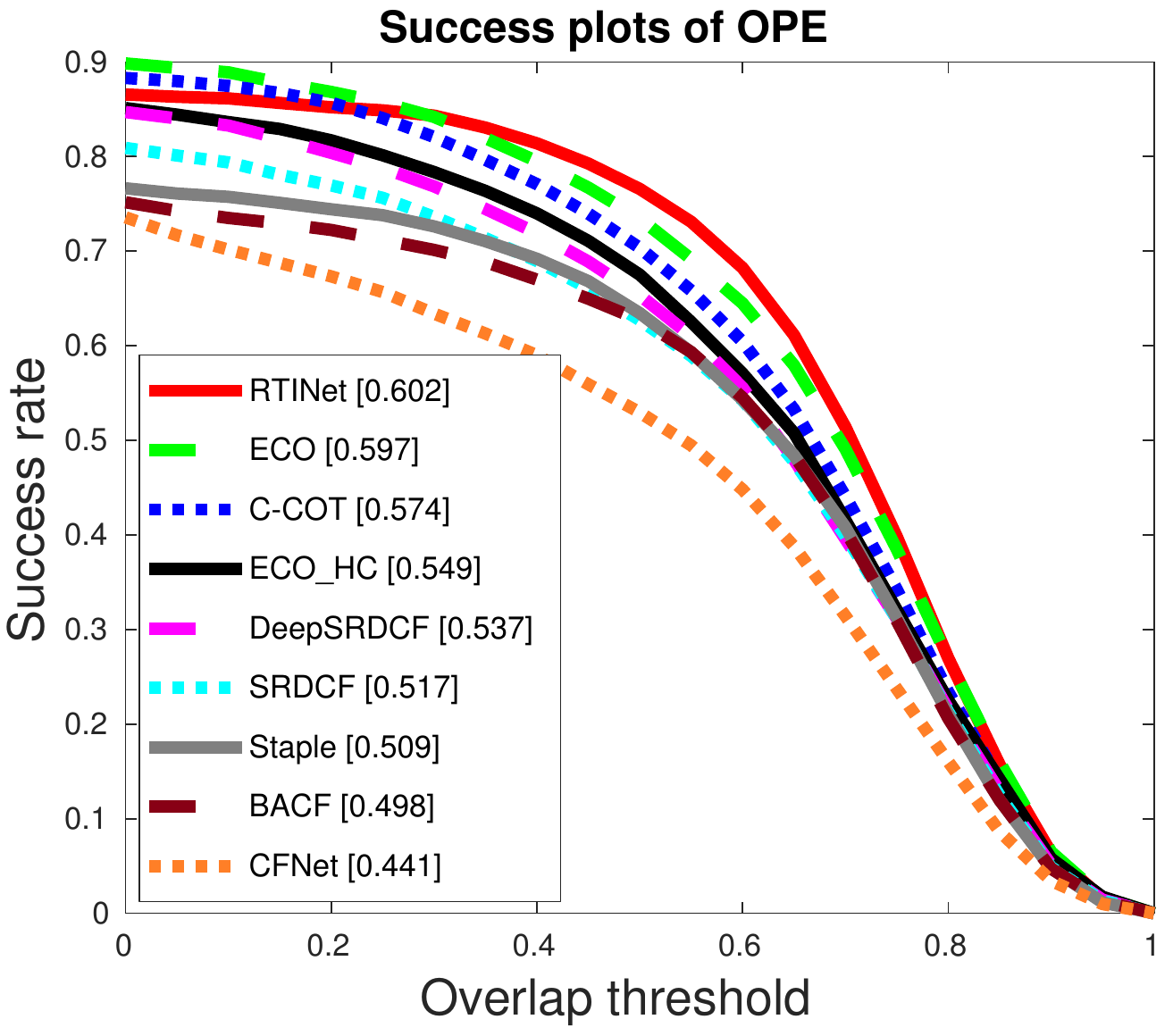}}
\caption{Overlap success plots of different trackers on the TB-50, OTB-2015 and TempleColor-128 datasets.}
\label{fig:auc}
\end{figure*}
%
%--------------------------------------------------------------------------------
\subsection{Comparison with the state-of-the-art methods}
We compare RTINet with several state-of-the-art trackers, including CF-based trackers
(i.e., ECO \cite{danelljan2017eco}, C-COT \cite{CCOT}, DeepSRDCF \cite{danelljan2015convolutional},
BACF \cite{BACF}, STAPLE-CA \cite{mueller2017context}) and
learning-based CNN trackers (i.e., MDNet \cite{MDNet}, MetaSDNet \cite{park2018meta},
MetaCREST \cite{park2018meta}, SiameseFC \cite{SiameseFC}, DSiamM \cite{guo2017learning} and SINT \cite{tao2016siamese}).
Note that all the results are obtained by using either the publicly available codes or the results provided by the authors for fair comparison.
Experiments are conducted on TB-50~\cite{wu2015object}, OTB-2015~\cite{wu2015object}, TempleColor-128~\cite{liang2015encoding} and VOT-2016~\cite{Kristan2016a}.
On the first three datasets, we follow the OPE protocol provided in~\cite{wu2015object} and
present the success plots ranked by the AUC scores.
On VOT-2016, we evaluate the trackers in terms of accuracy, robustness and expected average overlap (EAO).
%
%The accuracy measures the average overlap ratio between the predicted and ground truth bounding boxes.
%
%The robustness computes the average number of tracking failures over the entire sequence.
%
%And the EAO metric averages the no-reset overlap of a tracker on several short-term sequences.
%
%
{\flushleft {\bf OTB-2015 and TB-50.}}
Fig.~\ref{fig:auc}(a)(b) shows the success plots of the competing trackers the OTB-2015 and TB-50 benchmarks.
And the proposed RTINet is ranked in top-3 on the two datasets, achieves comparable performance with the top trackers such as ECO and MDNet~\cite{MDNet}.
Moreover, RTINet obtains an AUC score of {68.2}\% on OTB-2015, outperforming its counterparts CFNet and BACF by a margin of {11.4}\% and {6.7}\%, respectively.
In Fig.~\ref{fig:auc}, we also compare RTINet with the recently proposed Meta-Trackers \cite{park2018meta} (i.e., MetaSDNet and MetaCREST).
Again our RTINet performs better than both MetaSDNet and MetaCREST by the AUC score.
And even the rapid version RTINet-conv1 outperforms MetaCREST, and is comparable to MetaSDNet.
On the more challenging sequences in TB-50, our RTINet is still on par with the state-of-the-art ECO and ranks the second among the competing trackers.
Specifically, RTINet performs better than the other learning-based trackers,
including SiameseFC \cite{SiameseFC}, DSiamM \cite{guo2017learning} and
SINT \cite{tao2016siamese}, and surpasses its baseline CFNet \cite{CFNet} by {10.7}\%.
In comparison to CFNet and BACF, the superiority of RTINet can be ascribed to the incorporation of the advanced BACF model, and the joint learning of deep representation and truncated inference.
{Finally, we analyze the performance with respect to attributes. RTINet performs in top-3 on 6 of the 11 attributes and is on par with the state-of-the-arts on the other attributes. Detailed results are given in the supplementary materials. The results further validates the effectiveness of our proposed RTINet.} % the joint learning of features and CF model adaptation benefits to tackling various distractions
%
%For comprehensive comparison, we also analyze the performance with respect to different attributes of test videos in the OTB-2015 benchmark.
%
%RTINet performs in top three on 6 attributes as shown in Fig.~\ref{fig:auc_attributes} and is on par with the state-of-the-arts on other attributes.
%
%It confirms that the joint learning of features and CF model adaptation benefits to tackling various distractions. More results are provided in the supplementary.
%
%
%%%%%%%%%%%%%%%%%%%%%%%%%%%%%%%%%%
%\vspace{-2mm}
{\flushleft {\bf TempleColor-128.}}
Fig.~\ref{fig:auc}(c) shows the success plots on TempleColor-128.
RTINet performs favorably against ECO with an AUC score of 60.2\%,
and achieves significant improvements over BACF and C-COT, by a gain of 10.4\% and 2.8\%, respectively.
In particular, compared with its counterpart CFNet, RTINet improves the performance with a large margin of 16.1\%.
The results further demonstrate the effectiveness of joint representation and truncated inference learning.
%
%%%%%%%%%%%%%%%%%%%%%%%%%%%%%%%%%%
%\vspace{-2mm}
{\flushleft {\bf VOT2016.}}
Quantitative results on VOT2016 are also be presented in terms of accuracy,
robustness and EAO in Table~\ref{tab:vot2016}.
RTINet achieves promising performance and performs much
better than the BACF, SRDCF and DeepSRDCF both in terms of accuracy and robustness.
In particular, it obtains the {best} result on accuracy with a value of {0.57},
and performs the {third-best} on robustness and EAO. %with the failure rate of {1.07}.
%
%In terms of EAO, RTNet ranks the third-best among the state-of-the-arts.
%
%which further proves the effectiveness of the proposed joint learning framework based on BACF model.
%
It is worth noting that, RTINet performs favorably to ECO by accuracy but is inferior by robustness, which may be ascribed to that only the accuracy is considered in the training loss in Eqn. (\ref{eq:loss_RTI}) of RTINet.
%
%%%%%%%%%%%%%%%%%%%%%%%%%%%%%%%%%%%%%%%%%%%%%%%%%%%%%%%%%%%%%%%%%%%%%%
\begin{table*}[t]
%\scriptsize
  \caption{Comparison with the state-of-the-art trackers in terms of EAO, Robustness, and Accuracy on VOT-2016 dataset.}
  \label{tab:vot2016}
  \centering
 \resizebox{\textwidth}{!}{
  \begin{tabular}{lcccccccccc}
    \toprule
    %\multicolumn{2}{c}{Part}                   \\
%    \cmidrule{1-2}
    Trackers   &ECO &C-COT &DeepSRDCF &SRDCF &HCFT &Staple &BACF  &RTINet \\
    \midrule
    EAO        &{\color{red}{0.374}} &{\color{blue}{0.331}}   &0.276         &0.247     &0.220    &0.295      & 0.233        &       {\color{green}0.298}  \\
    Accuracy   &{\color{green}0.54}  &0.52      &0.51          &0.52      &0.47     &{\color{green}0.54}       & {\color{blue}0.56}       &               {\color{red}0.57} \\
    Robustness &{\color{red}0.72}  &{\color{blue}0.85}    &1.17          &1.50      &1.38     &1.35       &1.88     &               {\color{green}1.07} \\
    \bottomrule
  \end{tabular}
  }
\end{table*}
%%%%%%%%%%%%%%%%%%%%%%%%%%%%%%%%%%%%%%%%%%%%%%%%%%%%%%%%%%%%%%%%%%%%%%

\section{Conclusion}
\label{sec:conclusion}
This paper presents a RTINet framework for joint learning of deep representation and model adaptation in visual tracking.
We adopt the deep convolutional network for feature representation and integrate the CNN with advanced BACF tracker.
To solve the BACF in the CNN architecture, we design the model adaptation network as truncated inference by unrolling the ADMM optimization of the BACF model.
Moreover, a greedily stage-wise learning scheme is introduced for the joint learning of deep representation and truncated inference from the annotated video sequences.
Experimental results on three tracking benchmarks show that our RTINet tracker achieves favorable performance in comparison with the state-of-the-art trackers.
Besides, our rapid version of RTINet can run in real-time (24 fps) at a moderate sacrifice of accuracy.
By taken BACF as an example, our RTINet sheds some light on incorporating the advances in CF modeling for improving the performance of learning-based trackers, and thus deserves in-depth investigation in future work.
\flushleft{\bf Acknowledgement.} This work was supported in part by the National Natural Science Foundation of China under Grant No.s 61671182 and 61471146.

%\bibliographystyle{splncs04}
%\bibliography{RTINetTracker}
%

\end{document}